\documentclass[10pt,twocolumn,letterpaper]{article} 

\usepackage{cvpr}

\usepackage{times}
\usepackage{relsize}
\usepackage[T1]{fontenc} 
\usepackage[latin1]{inputenc} 
\usepackage[english]{babel}

\usepackage{epsfig}
\usepackage{graphicx}
\usepackage{wrapfig}
\usepackage[belowskip=0pt,aboveskip=1pt,font=small, labelfont=bf]{caption}
\usepackage{subcaption}
\usepackage{amsmath, amsthm, amssymb}
\usepackage{textcomp}
\usepackage{stmaryrd}
\usepackage{upgreek}
\usepackage{bm}
\usepackage{cases}
\usepackage{mathtools}
\usepackage{gensymb}

\usepackage[numbers,sort&compress,square,comma]{natbib}
\usepackage[pagebackref=true,breaklinks=true,letterpaper=true,colorlinks,bookmarks=false,citecolor=red]{hyperref}

\usepackage{algorithm}
\usepackage{algpseudocode}

\usepackage{multirow}
\usepackage{rotating}
\usepackage{booktabs}

\usepackage{enumitem}
\usepackage[olditem,oldenum]{paralist}

\usepackage{alltt}
\usepackage{listings}

\belowdisplayskip \abovedisplayskip

\newlength{\sectionReduceTop}
\newlength{\sectionReduceBot}
\newlength{\subsectionReduceTop}
\newlength{\subsectionReduceBot}
\newlength{\abstractReduceTop}
\newlength{\abstractReduceBot}
\newlength{\captionReduceTop}
\newlength{\captionReduceBot}
\newlength{\subsubsectionReduceTop}
\newlength{\subsubsectionReduceBot}

\newlength{\eqnReduceTop}
\newlength{\eqnReduceBot}

\newlength{\horSkip}
\newlength{\verSkip}

\newlength{\figureHeight}
\usepackage{mysymbols}

\usepackage{url}
\usepackage{xspace}
\usepackage{comment}
\usepackage{color}
\usepackage{afterpage}
\usepackage{pdfpages}
\usepackage{framed}
\usepackage{fancybox}

\usepackage{epstopdf}
\graphicspath{{Images/}}
\usepackage[stable]{footmisc}
\usepackage{tabulary}
\usepackage{tabularx}

\newcommand{\wb}{\mathbf{w}} 

\newcommand{\feat}{\bfgreek{phi}}

\cvprfinalcopy 

\def\cvprPaperID{241} 

\ifcvprfinal\fi

\begin{document}

\twocolumn[{
\title{VIP: Finding Important People in Images} 
\author{Clint Solomon Mathialagan\\
Virginia Tech\\
\and
Andrew C. Gallagher\\
Google Inc.\\
\and
Dhruv Batra\\
Virginia Tech\\
}

\maketitle
\ifcvprfinal\thispagestyle{empty}\fi
\vspace{-5mm}
\small Github: \url{https://github.com/mclint/vip}  \hfill
Datasets: \url{https://huggingface.co/matclint}
\vspace{5mm}

}]

\vspace{\abstractReduceTop}
\begin{abstract} 
\vspace{\abstractReduceBot}
People preserve memories of events such as birthdays, weddings, or vacations by capturing photos, 
often depicting groups of people. Invariably, some individuals in the image are more important than 
others given the context of the event. This paper analyzes the concept of the importance of individuals 
in group photographs. 
We address two specific questions -- 
Given an image, who are the most important individuals in it? 
Given multiple images of a person, which image depicts the person in the most important role? 
We introduce a measure of \emph{importance} of people in images and 
investigate the correlation between importance and visual saliency. 
We find that not only can we automatically predict the importance of people from purely visual cues, 
incorporating this predicted importance results in significant improvement 
in applications such as im2text (generating sentences that describe images of groups of people). 
\end{abstract} 
\vspace{\abstractReduceBot}


\vspace{\sectionReduceTop}
\section{Introduction}
\label{sec:intro}
\vspace{\sectionReduceBot}

\begin{figure}[t]
\centering

\includegraphics[width=1\columnwidth]{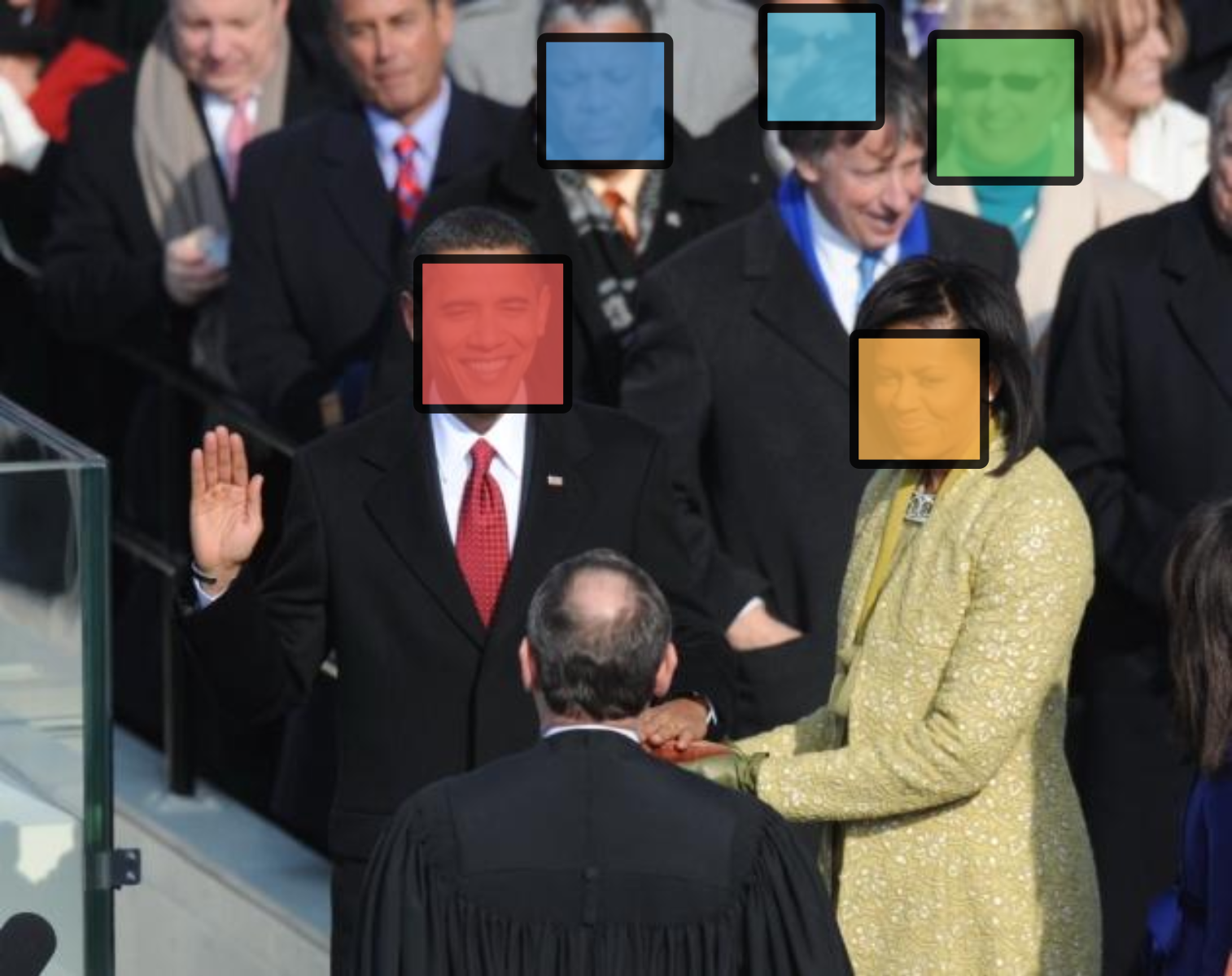}
%
%
\includegraphics[width=1\columnwidth]{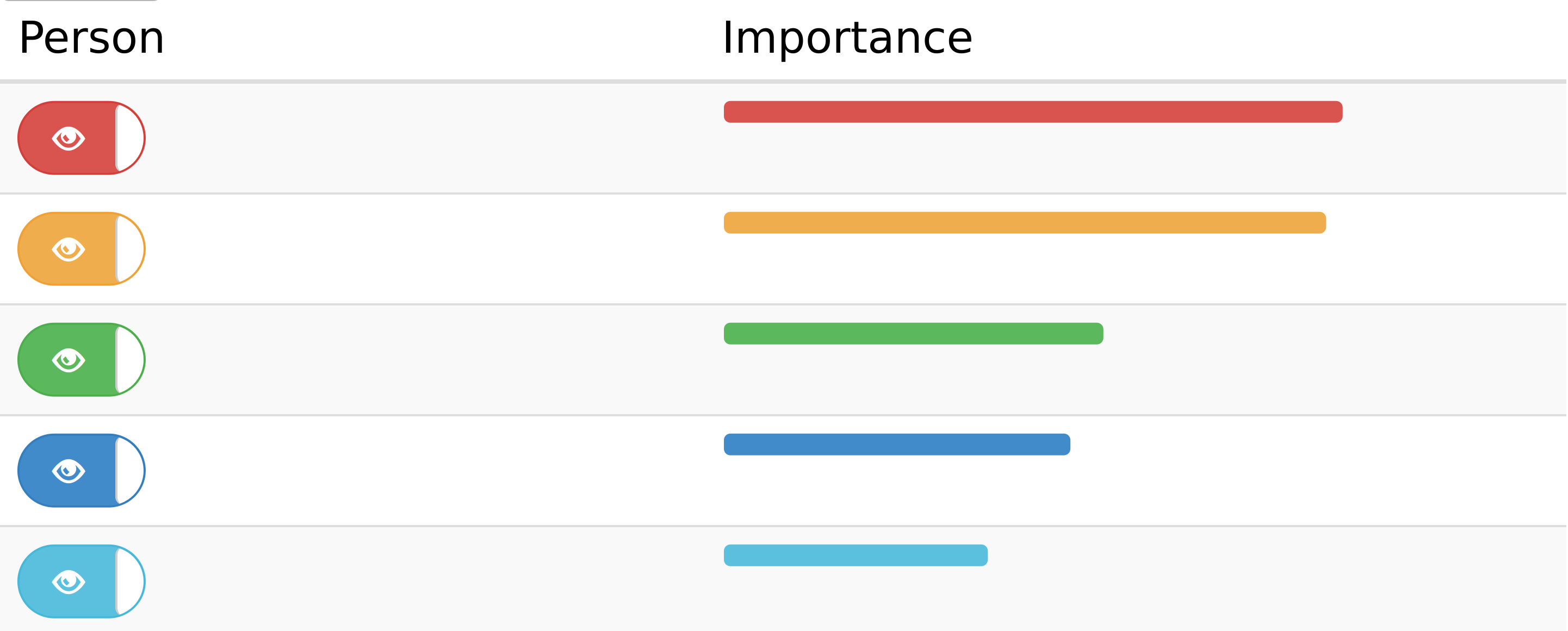}

\caption{Goal: Predict the importance of individuals in group photographs 
(without assuming knowledge about their identities).} 
\vspace{\captionReduceBot}
\vspace{-7pt}
\label{fig:teaser}
\end{figure}

When multiple people are present in a photograph, there is usually a story behind the situation 
that brought them together: a concert, a wedding, 
a presidential swearing-in ceremony (\figref{fig:teaser}), 
or just a gathering of a group of friends. In this story, not 
everyone plays an equal part. Some person(s) are the main character(s) and play a more 
central role. 

\begin{figure*}[t]
\centering

\begin{subfigure}[t]{0.32\textwidth}
\includegraphics[width=0.9\textwidth, height=33mm]{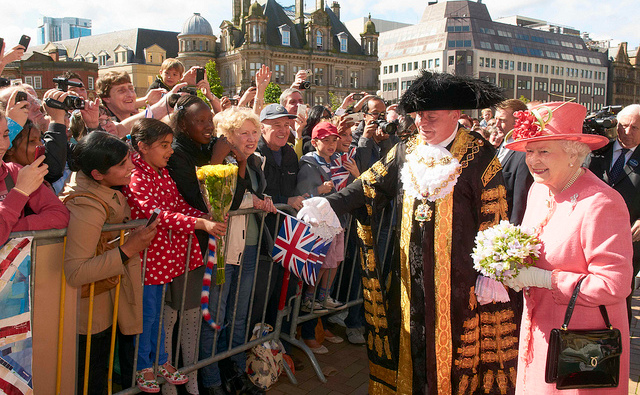}
\caption{Socially prominent people.}
\label{fig:intro1}
\end{subfigure}
\hfill
\begin{subfigure}[t]{0.32\textwidth}
\includegraphics[width=0.9\textwidth, height=33mm]{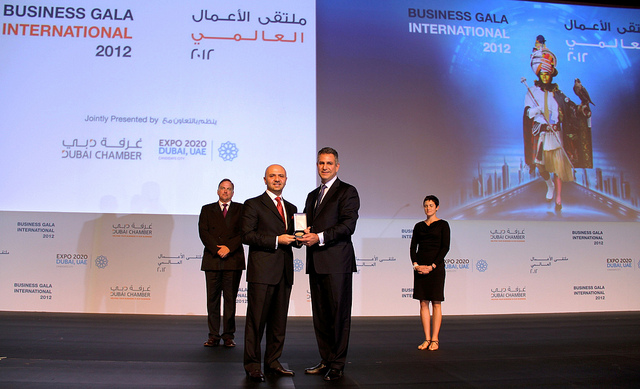}
\caption{Non-celebrities.}
\label{fig:intro2}
\end{subfigure}
\hfill
\begin{subfigure}[t]{0.32\textwidth}
\includegraphics[width=0.9\textwidth, height=33mm]{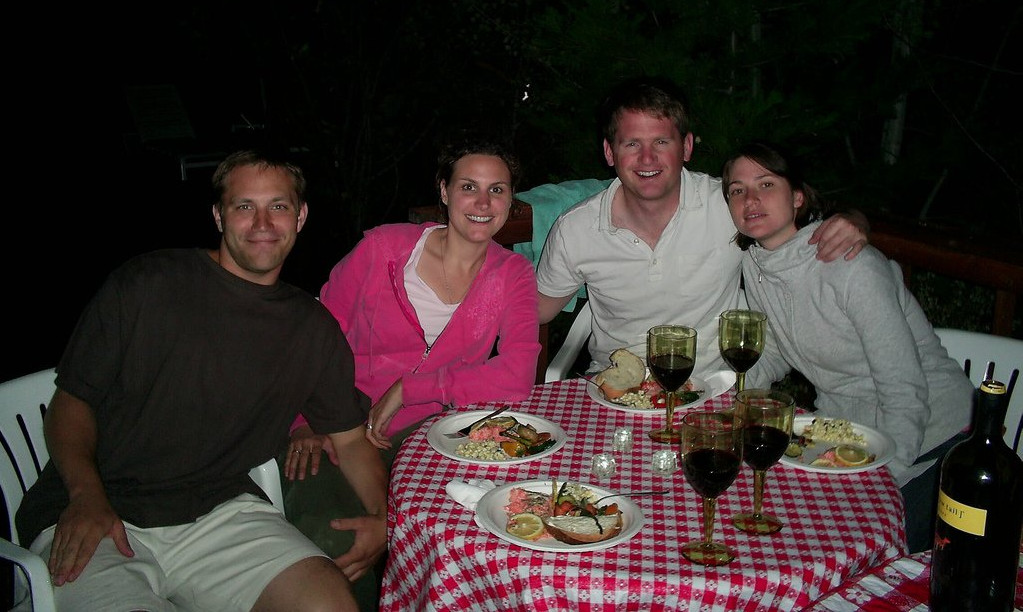}
\caption{Equally important people.}
\label{fig:intro3}
\end{subfigure}

\caption{Who are most important individuals in these pictures? 
(a) the couple (the British Queen and the Lord Mayor);  
(b) the person giving the award and the person receiving it play the main role; 
(c) everyone seems to be nearly equally important. 
Humans have a remarkable ability to understand social roles and identify important players, 
even without knowing identities of 
the people in the images. 
} \vspace{-10pt}
\label{fig:introExamples}
\end{figure*}

Consider the picture in \figref{fig:intro1}. Here, the important 
characters are the couple who appear to be the British Queen and the Lord Mayor. 
Notice that their identities and social status play a role in establishing their positions as the 
key characters in that image. 
However, 
it is clear that even someone unfamiliar with the oddities and eccentricities of the British Monarchy,  
who simply views this as a picture of an elderly woman and a gentleman in costume  
receiving attention from a crowd, would consider 
those two to be central characters in that scene. 

\figref{fig:intro2} shows an example with people who do not appear to be celebrities. 
We can see that two people in foreground are clearly the focus of attention, and 
two others in the background are not. 
\figref{fig:intro3} shows a common group photograph, 
where everyone is nearly equally important.
It is clear that even without recognizing the identities of people, we as humans 
have a remarkable ability to understand social roles and identify important players. 

\textbf{Goal and Overview.} 
The goal of our work is to \emph{automatically predict the importance of individuals in group photographs}. 
In order to keep our approach general and applicable to any input image, 
we focus purely on visual cues available in the image, and do not assume identification of 
the individuals. Thus, we do not use social prominence cues. For example, in \figref{fig:intro1}, 
we want an algorithm that identifies the elderly woman and the gentleman as the top-2 most important 
people that image 
without utilizing the knowledge that the elderly woman is the British Queen. 

\textbf{What is Importance?}
In defining importance, 
we can consider the perspective of three parties (which may disagree): 
%
\begin{compactitem}

\item \textbf{the photographer}, who presumably intended to capture some subset of people, 
and perhaps had no choice but to capture others; 
\item \textbf{the subjects}, who presumably arranged themselves following social inter-personal rules;  and
\item \textbf{neutral third-party human observers}, who may be unfamiliar with the subjects of the photo 
and the photographer's intent, but may still agree on the (relative) importance of people. 

\end{compactitem}
%
Navigating this landscape of perspectives 
involves many complex social relationships: the social 
status of each person in the image (an award winner, a speaker, the President), and 
the social biases of the photographer and the viewer (\eg, gender or racial biases); 
many of these can not be easily mined from the photo itself. 
At its core, the question itself is subjective: 
if the British Queen ``photo-bombs'' while you are taking a picture of your friend, 
is she still the most important person in that photo? 

In this work, to establish a quantitative protocol, 
we rely on the wisdom of the crowd to 
estimate the ``ground-truth'' importance of a person in an image. 
We found the design of the annotation task and the interface to be particularly important, 
and discuss these details in the paper. 


\textbf{Applications.} 
A number of applications can benefit from knowing the importance of people. 
Algorithms for im2text (generating sentences that describe an image) can be made more human-like if they 
describe only the important people in the image and ignore unimportant ones. 
Photo cropping algorithms can do ``smart-cropping''  of images of people by keeping 
only the important people. 
Social networking sites and 
image search applications can benefit from improving the ranking of photos where 
the queried person is important. 

\textbf{Contributions.} 
This paper makes the following contributions. 
First, we learn a model for predicting importance of individuals in photos based on 
a variety of features that capture the pose and arrangement of the people. 
Second, we collect two importance datasets that serve to evaluate our approach, and will 
be broadly useful to others in the community studying related problems. 
Finally, we show that we can automatically predict the importance of people with high accuracy, 
and incorporating this predicted importance in applications such as im2text leads to significant improvement.
Despite the naturalness of the task, to the best of our knowledge, this is the 
first paper to directly infer the importance of individuals in the context of a single group image.

\vspace{\sectionReduceTop}
\section{Related Work}
\label{sec:related}
\vspace{\sectionReduceBot}


\textbf{General Object Importance.} 
Our work is related to a number of previous works \cite{spainImportance, graumanImportance, bergImp} 
that study the importance of generic object categories. 
Berg~\etal~\cite{bergImp} define importance of an object as 
the likelihood that it will be mentioned in a sentence written by a person 
describing the image. 
The key distinction between their work and ours is that they study the problems at a 
category level (``are people more important than dogs?''), 
while we study it at an instance level,  
restricted to instances of people 
(``is person A more important than person B in this image?''). 
One result from \cite{bergImp} is that `person' generally tends to 
be the most important category. 
Differentiating between the importance of different individuals in an image  
produces a more fine-grained understanding of the image.
Le~\etal~\cite{impLargeNews} consider people who have appeared repeatedly in a certain time period from large news video databases to be important.
Lee~\etal~\cite{impVideo} study importance of objects (including people) in  
egocentric videos, where important things are those with which the camera wearer has significant interaction. 
In our work, we focus on a single image, and do not assume access to user-attention cues. 

\textbf{Visual Saliency.} 
A number of works~\cite{ittiSal, liuSal, harelSal} have studied visual saliency 
-- identifying which parts of an image draw viewer attention. 
Humans tend to be a naturally salient content in images. 
Jiang~\etal~\cite{jiangCrowd} study visual saliency in group photographs and crowded scenes. 
Their objective is to build a visual saliency model that takes into account the presence 
of faces in the image. Although they study the same content as our work (group photographs), the goals 
of the two are different -- saliency vs importance. 
At a high level, saliency is about what draws the viewer's attention; importance is a higher-level 
concept about social roles. 
We conduct extensive human studies and discuss this 
comparison in the paper. Saliency is correlated to, but not identical to importance. 
People in photos may be salient but not important, important but not salient, 
both, and neither. 

\textbf{Understanding Group Photos.} 
Our work is related to a line of work in Computer Vision studying photographs of groups of people 
\cite{andyGroups, amirPeople, andyGroups1, andyGroups2, andyGroups3}, 
addressing issues such structural formation and attributes of groups. 
Li~\etal~\cite{congAesthetics} predict the aesthetics of a group photo. 
If the measure is below a threshold, photo cropping is suggested by eliminating 
unimportant faces and regions that do not seem to fit in with the general structure of the group. 
While their goal is closely related to ours, 
they study \emph{aesthetics}, not importance. 
To the best of our knowledge, this is the first work to predict importance of individuals in a group photo. 

\begin{figure*}[t]
\centering

\begin{subfigure}[t]{0.49\textwidth}
\includegraphics[width=1\textwidth, height=48mm]{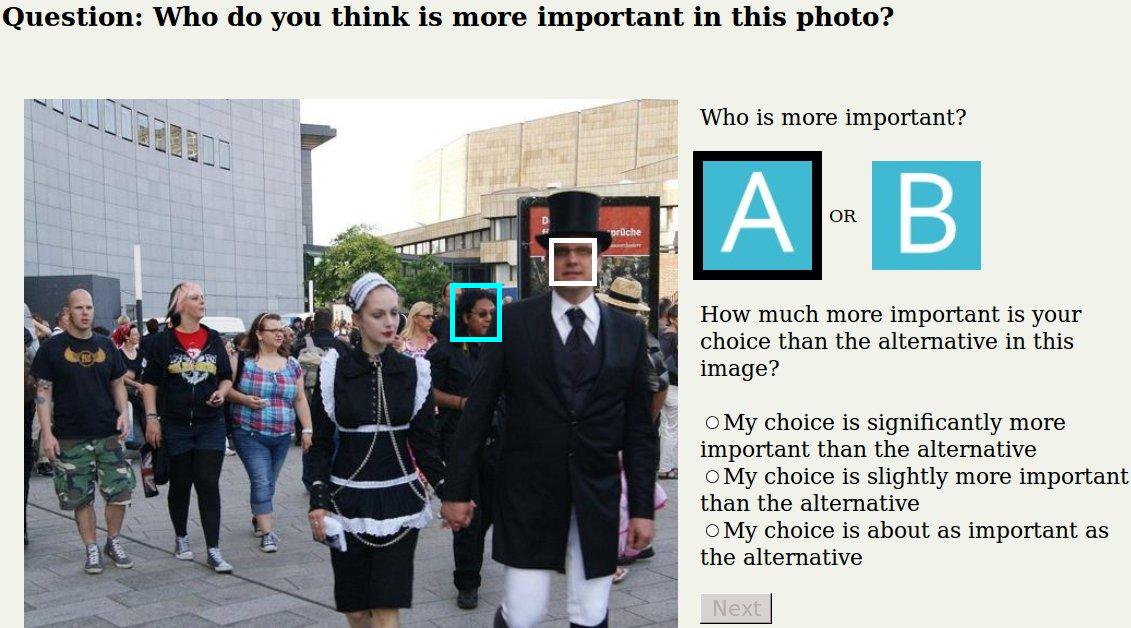}%
\caption{Image-Level annotation interface.}
\label{fig:interface1}
\end{subfigure}
\hfill
\begin{subfigure}[t]{0.49\textwidth}
\includegraphics[width=1\textwidth, height=48mm]{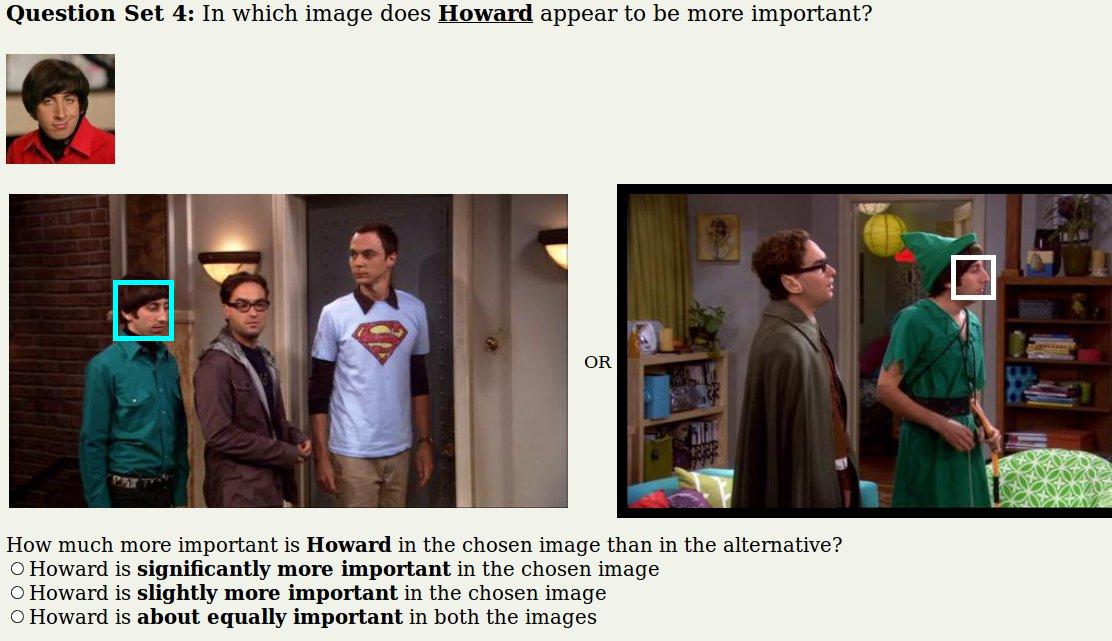}%
\caption{Corpus-Level annotation interface.}
\label{fig:interface2}
\end{subfigure}

\caption{Annotation Interfaces used with MTurk: 
(a) Image-Level: Hovering over a button (A or B) highlights the person associated with it 
(b) Corpus-Level: Hovering over a frame shows the where the person is located in the frame.} \vspace{-10pt}
\label{fig:annotInterfaces}
\end{figure*}

\vspace{\sectionReduceTop}
\section{Approach}
\label{sec:approach}
\vspace{\sectionReduceBot}

Recall that our goal is to model and predict the importance of people in images. 
We model importance in two ways: 
\begin{compactitem}
\item \textbf{Image-Level Importance:}
``Given an image, who is the most important individual?''
This reasoning is local to the image in question. The 
objective is to predict an importance score for each person in the image. 

\item \textbf{Corpus-Level Importance:}
``Given multiple images, in which image is a specific person most important?'' 
This reasoning is across a corpus of photos (each containing a person of interest), 
and the objective is to assign an importance score to each image. 

\end{compactitem}

\vspace{\subsectionReduceTop}
\subsection{Dataset Collection}
\vspace{\subsectionReduceBot}

For each setting, we curated and annotated a dataset. 

\textbf{Image-Level Dataset.} \cite{im_dataset}
In this setting, we need a dataset of images containing at least three people with varying levels 
of importance. While the `Images of Groups' dataset \cite{andyGroups} initially seems like a good candidate, 
it is not suitable for studying importance because there is little change in \emph{relative} importance
-- most images are posed group photos where 
everyone is nearly equally important (\eg \figref{fig:intro3}).  

We collected a dataset of 200 images by mining Flickr for images (with appropriate licenses) 
using search queries such as ``people+events'', ``gathering", \etc. Each image has three 
or more people in varying levels of importance. 
In order to automatically predict the importance of individuals in the image, they need to be detected first. 
For the scope of this work, we assume face detection to be a solved problem. 
Specifically, the images in our dataset were first run through a face detection API \cite{skybiometry}, 
which has a fairly low false positive rate. 
Missing faces and heads were then annotated manually. 
There are in total 1315 annotated people in the dataset, with $\sim$6.5 persons per image on average. 
Example images are shown throughout the paper and 
the dataset is publicly available 
from the project webpage~\cite{project_page}. 

\textbf{Corpus-Level Dataset.} \cite{corpus_dataset}
In this setting, we need a dataset that has multiple pictures of the same person; 
and multiple sets of such photos.  
The ideal source for such a dataset are 
social networking sites. However, privacy concerns hinder the annotation of these images via crowdsourcing. 
TV series, on the other hand, have multiple frames with the same people and 
are good sources to obtain such a dataset. Since temporally-close 
frames tend to be visually similar, these videos should be properly sampled to get diverse images. 

The personID dataset by Tapaswi~\etal~\cite{tapaswiPerson} contains face track 
annotations (with character identification) for the first six episodes of the `Big Bang Theory' TV series. 
The track annotation of a person gives the coordinates of face bounding boxes for 
the person in every frame. 
By selecting only one frame from each track of a character, one can get diverse frames 
for that character from the same episode. From each track, we selected the frame that 
has the most people. 
Some selected frames have only one person in them, but that is acceptable since the 
task is to pick the most important frame for a person. 
In this manner, a distinct set of frames was obtained for each of the five main characters 
in each episode.

\vspace{\subsectionReduceTop}
\subsection{Importance Annotation}
\vspace{\subsectionReduceBot}

We collected ground-truth importance in both datasets via 
Amazon Mechanical Turk (AMT). 
We conducted pilot experiments to identify the best way to 
annotate these datasets, and pose the question of importance. 
We found that when subjects were posed an absolute question  
``Please mark the important people in this image,''
they found the task difficult. 
Turkers commented that they had to redefine their notion of importance for each new image, 
making consistency difficult. Indeed, we observed low inter-human agreement, 
with some workers selecting everyone in the image as important, 
and others selecting no more than one person. 

To overcome these inconsistencies, we redesigned the tasks to be relative 
(details next). 
This made each task 
simpler, and the annotations more consistent. 

\textbf{Image-Level Importance Annotation.} 
From each image in the image-level dataset, random pairs of faces were 
selected to produce a set of 1078 pairs. These pairs cover $91.82\%$ of the 
total faces in these images. 
For each selected pair, ten AMT workers were asked to pick 
the more important of the two. 
The interface is shown in \figref{fig:interface1}, and an HTML version 
is available from the project webpage \cite{project_page}. 
In addition to clicking on the more important face, the workers were also asked to report 
magnitude of the difference in importance between the two people: 
\emph{significantly different}, \emph{slightly different} and \emph{almost same}. 
This forms a three-tier scoring system as depicted in Table \ref{tab:threePointSystem}.

\begin{table}[h]
\centering
\begin{tabular}{@{\extracolsep{\fill}}p{4.5cm}|cc@{}}
\toprule
Turker selection: A is  & A's score & B's score \\
\midrule
{\small \emph{significantly more} important than B} & 1.00 & 0.00 \\
{\small \emph{slightly more} important than B} & 0.75 & 0.25 \\
{\small \emph{about as} important as B} & 0.50 & 0.50\\
\bottomrule
\end{tabular}
\caption{Converting pairwise annotations to importance scores.}
\label{tab:threePointSystem}
\vspace{-15pt}
\end{table}

For each annotated pair of faces $(p_i,p_j)$ 
the 
relative importance scores $s_i$ and $s_j$ range from 0 to +1, 
and indicates the relative difference in importance between $p_i$ and $p_j$. 
Note that $s_i$ and $s_j$ are not absolute, as they are not calibrated for comparison 
to another person, say $p_k$ from another pair.

\textbf{Corpus-Level Importance Annotation.}
From the corpus-level dataset, approximately 1000 pairs of 
frames were selected. Each pair contains frames depicting the same person but 
from different episodes. This ensures that the pairs do not contain similar 
looking images. AMT workers were shown a pair of frames for a 
character and asked to pick the frame where the character appears to be more 
important. The interface used is as shown in \figref{fig:interface2}, and an HTML version 
is available from the project webpage \cite{project_page}. 

Similar to the previous setting, 
workers were asked to pick a frame and indicate 
the magnitude of difference in importance of the character. 
These qualitative magnitude choices were converted into scores as in shown 
Table \ref{tab:threePointSystem}. 

Table \ref{tab:pairDistributions} shows a breakdown of both datasets 
along the magnitude of differences in importance. 
We note some interesting similarities and differences. 
Both datasets have nearly the same percentage of pairs that are `almost-same'. 
The instance-level dataset has many more pairs in the `significantly-more' category 
than the corpus-level dataset. 
This is because in a TV series dataset, the characters 
in a scene are usually playing some sort of a role in the scene, unlike typical consumer photographs 
that tend to contain many people in the background. 
Overall, both datasets contain a good mix of the three categories. 

%

\begin{table}[h]
\centering
\begin{tabular}{@{\extracolsep{\fill}}p{3cm}|cc@{\extracolsep{\fill}}}
\toprule
Pair category & Image-Level & Corpus-Level \\
\midrule
{\small significantly-more} & {\small $32.65\%$} & {\small $18.30\%$} \\
{\small slightly-more} & {\small $20.41\%$} & {\small $39.70\%$} \\
{\small almost-same} & {\small $46.94\%$} & {\small $42.00\%$} \\
\bottomrule
\end{tabular}
\caption{Distribution of Pairs in the Datasets.}
\label{tab:pairDistributions}
\vspace{-5pt}
\end{table}

\vspace{\subsectionReduceTop}
\subsection{Importance Model}
\vspace{\subsectionReduceBot}

We now formulate a relative importance prediction model that is applicable to both tasks: 
image-level and corpus-level. 
As we can see from the dataset characteristics in Table \ref{tab:pairDistributions}, 
our model should not only be able to say which person 
is more important, but also predict the relative strengths between pairs of people/images. 
Thus, we formulate this as a regression problem. 
Specifically, given a pair of people $(p_i,p_j)$ (coming from the same or different images) 
with scores $s_i, s_j$, the objective is to build a model $M$ that 
regresses to the difference in ground truth importance score: 
\begin{align}
M(p_i,p_j) \approx S_{i} - S_{j} \label{eqn1}
\end{align}
We use a linear model: $M(p_i,p_j) = \wb^\intercal \feat(p_i,p_j)$, 
where $\feat(p_i,p_j)$ are the features extracted for this pair, 
and $\wb$ are the regressor weights. 
We use $\nu$-Support Vector Regression to learn these weights. 
%
%
%
Our pairwise feature $\feat(p_i,p_j)$ are composed from features extracted for individual people 
$\feat(p_i)$ and $\feat(p_j)$. 
In our preliminary experiments, 
we compared two ways of composing these individual face features 
-- using difference of features $\feat(p_i,p_j) = \feat(p_i) - \feat(p_j)$; and concatenating 
the two individual features $\feat(p_i,p_j) = [\feat(p_i); \feat(p_j)]$. We found difference of features to work 
better, and all results in this paper are reported with that. 

\vspace{\subsectionReduceTop}
\subsection{Person Features}
\vspace{\subsectionReduceBot}

We now describe the features we used to assess importance of a person. 
Recall that we assume that faces in the images have been detected (by running a standard 
face detector). 



\textbf{Distance Features.}
We use a number of different ways to capture distances between faces in the image. 

Photographers often frame their subjects. 
In fact, a number of previous works~\cite{ittiCenter, ittiCenter1, tatlerCenter}
have reported a ``center bias'' -- 
the objects or people closest to the center tend to be the most important. 
Thus, we first scale the image to a size of (1, 1), and 
compute two distance features: \\ 
Distance from center: The distance from the center of 
the face bounding box 
to the center of the image (0.5, 0.5). 

Weighted distance from center: The previous feature divided by the largest 
dimension of the face box, so that larger faces are not 
considered to be farther from the center. 

We compute two more features to 
capture how far a person is from the center of a group: \\
Normalized distance from centroid: First, we find the centroid 
of all the center points of the face boxes. Then, we compute the distance of a face to this centroid. 

Normalized distance from weighted centroid: 
Here, the centroid is calculated as the weighted average of center 
points of faces, the weight of a face being the ratio of the area of the head to the 
total area of faces in the image.

\textbf{Scale.}
Large faces in the image often correspond to people who are closer to the camera, and 
perhaps more important. 
This feature is a ratio of the area of the face bounding box to the 
the area of the image.

\textbf{Sharpness.}
Photographers often use a narrow depth-of-field to keep the indented subjects in focus, 
while blurring the background. 
In order to capture this phenomenon, we compute a sharpness feature in every face. 
We apply a Sobel filter on the image and 
compute the the sum of the gradient energy in a face bounding box, 
normalized by the sum of the gradient energy in all the bounding boxes in the image.

\textbf{Face Pose Features.}
The facial pose of a person can be a good indicator of their importance, because important 
people often tend to be looking directly at the camera. 

DPM face pose features: We resize the face bounding box patch from the image to 128$\times$128 
pixels, and run the face pose and landmark estimation algorithm of 
Zhu~\etal~\cite{zhuFace}. Note that \cite{zhuFace} is mixture model where 
each component corresponds to a 
an the angle of orientation of the face, in the range of 
-90$^{\circ}$ to +90$^{\circ}$ in steps of 15$^{\circ}$. 
Our pose feature is this component id, 
which can range from 1 to 13. We also use a 13-dimensional indicator feature that 
has a 1 in the component with maximum score and zeros elsewhere. 

Aspect ratio: We also use the aspect ratio of the face bounding box is as a feature. 
While the aspect ratio of a face is typically 1:1, this ratio can differentiate 
between some head poses such as frontal and lateral poses.

DPM face pose difference: 
It is often useful to know where a crowd is looking and where a particular person is looking. 
To capture this pose difference between a person and others, 
we compute the pose of the person 
subtracted by the average pose of every other person in the image, as a feature. 

\textbf{Face Occlusion.}
Unimportant people are often occluded by others in the photo. Thus, we extract features 
to indicate whether a face might be occluded. 

DPM face scores: We use the difficulty in being detected as a proxy for occlusion. 
Specifically, we use scores for each the 13 components in the face detection 
model of~\cite{zhuFace} as a feature. We also use the score of the dominant component. 

Face detection success: This is a binary feature 
indicating whether the face detection API~\cite{skybiometry} we used was successful in detection 
the face, or whether it required human annotation. The API 
achieved a nearly zero false positive rate on our dataset. Thus, this feature 
served a proxy for occlusion since that is where the API usually failed. 
Note that this feature requires human inspection and would not be available to a fully-automatic approach. 
An online demo of our system available at \cite{cloudcv_vip, cloudcv} does not use this feature. 


In total, we extracted 45 dimensional features for every face.

\vspace{\sectionReduceTop}
\section{Results}
\label{sec:results}
\vspace{\sectionReduceBot}

For both datasets, we perform cross-validation on the annotated pairs. 
Specifically, we split the annotated pairs into 10 folds. 
We train the SVRs on 8 folds, pick hyper-parameters ($C$ in the SVR) on 1 validation 
fold, and make predictions on 1 test fold. 
This process is repeated for each test fold, 
and we report the average across all 10 test folds. 

\textbf{Baselines.} 
We compare our proposed approach to three natural baselines: 
center, scale, and sharpness baselines, 
where the person closer to the center, larger, or more in focus (respectively) 
is considered more important. The center baseline uses the weighted distance from center 
which not only gives priority to distance from the center but also to the size of the face. 
In order to measure how well a saliency detector performs on the importance prediction task, 
we used the 
method of Harel~\etal~\cite{graphSal, graphSalImpl} 
to produce saliency maps and computed the fraction of saliency intensities inside each face as a 
measure of its importance.

We measure inter-human agreement in a leave-one-human-out manner. In each iteration, 
responses of nine workers are averaged to get the ground-truth, and the response of the tenth 
worker is evaluated as the human response. 
This is then repeated for all ten human responses and the average is reported as inter-human 
agreement. 
In order to keep all automatic methods comparable to these inter-human results, we train 
all methods ten times, once for each leave-one-human-out ground-truth, and report the average results.

\textbf{Metrics.} 
We use mean squared error to measure the performance of our relative importance regressors. 
In addition, we convert the regressor output into binary classification by thresholding against zero. 
For each pair of faces $(p_i, p_j)$, we use a weighted classification accuracy measure, 
where the weight is 
the ground-truth importance score of the more important of the two, \ie $\max\{s_i,s_j\}$. 
Notice that this metric cares about the correct classification of 
`significantly-more' pairs more than the other pairs, which is natural. 

\textbf{Image-Level Importance Results.}
Table \ref{tab:results1} shows the results for different methods. 
We can see that the best baseline achieves 89.55\% weighted accuracy, 
whereas our approach achieves 92.72\%. Overall, we achieve an improvement 
of $3.17\%$ ($3.54\%$ relative improvement).
The mean squared error for our SVR is 0.1489.

\begin{table}[h]
\centering
\begin{tabular}{@{\extracolsep{\fill}}p{3cm}|c@{\extracolsep{\fill}}}
\toprule
Method & Weighted accuracy  \\
\midrule
{\small Inter-human agreement} & {\small $96.68 \pm 0.40\%$} \\
{\small Our approach} & {\small $\mathbf{92.72} \pm 0.93\%$} \\
\midrule
{\small Saliency detector} & {\small $83.52 \pm 1.29\%$} \\
\midrule
{\small Center baseline} & {\small $89.55 \pm 1.12\%$} \\
{\small Scale baseline}  & {\small $88.46 \pm 1.13\%$} \\
{\small Sharpness baseline}  & {\small $87.45 \pm 1.20\%$} \\
\bottomrule
\end{tabular}
\caption{Image-Level: Performance compared to baselines.}
\label{tab:results1}
\end{table}

Table \ref{tab:category1} show a break-down of the accuracies into the three categories of annotations. 
We can see that our approach outperforms the strongest baseline (Center) in every category, 
and the largest difference happens in the `significantly-more' category, which is quite useful. 

\begin{table}[h]
\centering
\begin{tabular}{@{\extracolsep{\fill}}p{2.5cm}|ccc@{\extracolsep{\fill}}}
\toprule
Pair category & Ours & C-Baseline & Improvement \\
\midrule
{\small significantly-more} & {\small $94.66\%$} & {\small $86.65\%$} & {\small $8.01\%$} \\
{\small slightly-more} & {\small $78.80\%$} & {\small $76.36\%$} & {\small $2.44\%$} \\
{\small almost-same} & {\small $55.98\%$} & {\small $52.96\%$} &  {\small $3.02\%$} \\


\bottomrule
\end{tabular}
\caption{Image-Level: Category-wise distribution of our predictions compared to Center baseline.}
\label{tab:category1}
\end{table}

\begin{figure*}[t]
\includegraphics[width=1\textwidth]{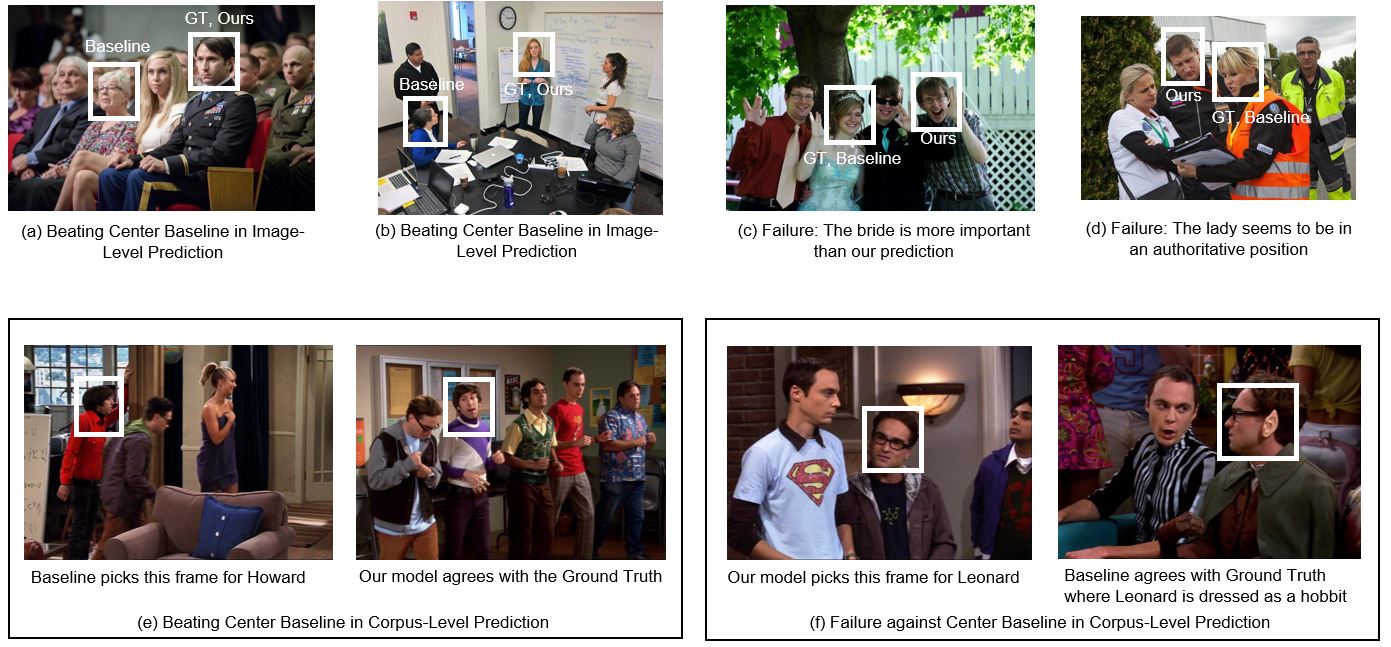}
\caption{Some results: (a)(b)(c)(d) for Image-Level prediction and (e)(f) for Corpus-Level prediction}
\label{fig:pairExamples}
\end{figure*}

\figref{fig:pairExamples} shows some qualitative results. 
We can see that individual features such center, sharpness, scale, and face occlusion 
help in different cases. 
In 3(c), the woman in blue is judged to be the most important, presumably because she is a bride. 
Unfortunately, our approach does not contain any features that can pick up on such social roles. 

\textbf{Corpus-Level Importance Results.}
Table \ref{tab:results2} shows the results for the corpus-level experiments. 
Interestingly, the strongest baseline in this setting is sharpness, rather than the center. 
This makes sense since the dataset is derived from professional videos; 
the important person is more likely to in focus compared to others.
Our approach outperforms all baselines, with an improvement of $4.18\%$ ($4.72\%$ relative improvement). 
The mean squared error is 0.1078.

\begin{table}[h]
\centering
\begin{tabular}{@{\extracolsep{\fill}}p{3cm}|c@{\extracolsep{\fill}}}
\toprule
Method & Weighted accuracy  \\
\midrule
{\small Inter-human agreement} & {\small $92.80 \pm 0.68\%$} \\
{\small Our approach} & {\small $\mathbf{92.70} \pm 0.77\%$} \\
\midrule
{\small Saliency detector} & {\small $89.26 \pm 1.20\%$} \\
\midrule
{\small Center baseline} & {\small $86.07 \pm 1.08\%$} \\
{\small Scale baseline}  & {\small $85.86 \pm 0.99\%$} \\
{\small Sharpness baseline}  & {\small $88.52 \pm 1.13\%$} \\
\bottomrule
\end{tabular}
\caption{Corpus-Level: Performance compared to baselines.}
\label{tab:results2}
\vspace{-5pt}
\end{table}

Table \ref{tab:category2} shows the category breakdown. While our method does extremely well with 
`significantly-more' pairs, it performs poorly in the `almost-same' category.

\begin{table}[t]
\centering
\begin{tabular}{@{\extracolsep{\fill}}p{2.5cm}|ccc@{\extracolsep{\fill}}}
\toprule
Pair category & Ours & S-Baseline & Improvement \\
\midrule
{\small significantly-more} & {\small $96.35\%$} & {\small $68.33\%$} & {\small $28.02\%$} \\
{\small slightly-more} & {\small $83.18\%$} & {\small $71.82\%$} & {\small $11.36\%$} \\
{\small almost-same} & {\small $58.36\%$} & {\small $69.93\%$} & {\small $-11.57\%$} \\


\bottomrule
\end{tabular}
\caption{Corpus-Level: Category-wise distribution of our predictions compared to Sharpness baseline.}
\label{tab:category2}
\vspace{-10pt}
\end{table}

\begin{table}[h]
\centering
\begin{tabular}{@{\extracolsep{\fill}}p{3cm}|cc@{\extracolsep{\fill}}}
\toprule
Features & Image-Level & Corpus-Level \\
\midrule
{\small All} & {\small $92.72 \pm 0.93\%$} & {\small $92.70 \pm 0.77\%$}\\
\midrule
{\small Without center} & {\small $91.25 \pm 0.95\%$} & {\small $92.41 \pm 0.71\%$}\\
{\small Without scale} & {\small $92.86 \pm 0.99\%$} & {\small $92.43 \pm 0.86\%$} \\
{\small Without sharpness} & {\small $92.22 \pm 1.10\%$} & {\small $91.52 \pm 1.31\%$} \\
{\small Only scale, center and sharpness}  & {\small $89.53 \pm 1.13\%$} & {\small $90.54 \pm 1.81\%$} \\
\bottomrule
\end{tabular}
\caption{Feature Ablation: Image-Level and Corpus-Level.}
\label{tab:ablation}
\end{table}

\figref{fig:pairExamples} also shows qualitative results for corpus experiments. 
Table \ref{tab:ablation} reports results from an ablation study, which shows the impact of the features 
on the final performance. 

\vspace{\sectionReduceTop}
\section{Importance vs Saliency}
\label{sec:saliency}
\vspace{\sectionReduceBot}

\begin{figure*}[t]
\includegraphics[width=1\textwidth]{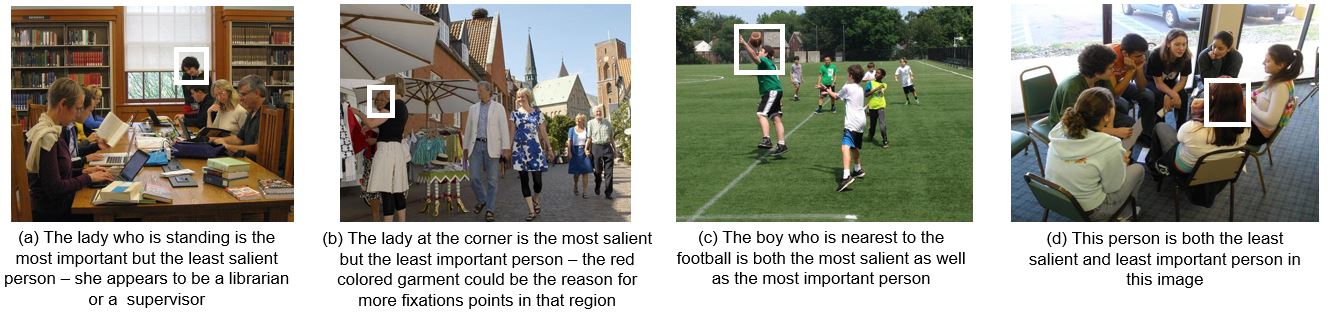}
\caption{Examples showing the relationship between visual saliency and person importance}
\label{fig:salExamples}
\vspace{-10pt}
\end{figure*}

Now that we know we can effectively predict importance, it is worth investigating how importance 
compares with visual saliency. 
At a high level, saliency studies 
what draws a viewer's attention in an image. 
Eye-gaze tracking systems are often used to track human eye fixations 
and estimate pixel-level saliency maps for an image. 
Saliency is potentially different from importance because saliency 
is controlled by low-level human visual processing, while importance involves 
understanding more nuanced social and semantic cues. 
However, as concluded by \cite{ittiSal}, important objects 
stand out in an image and 
are typically salient. 

We have already seen in Tables \ref{tab:results1}, \ref{tab:results2} that saliency detectors perform worse than 
baselines in the image-level task and worse than our model in the corpus-level task respectively. 

So how much does the salience of a face correlate with the importance of the person? 
We answer this question via the dataset collected by Jiang~\etal~\cite{jiangCrowd}  
to study saliency in group photos and crowded scenes. 
The dataset contains eye fixation annotations and face bounding boxes. 
For the purpose of this evaluation, we reduced the dataset to images with a minimum 
of 3 and maximum of 7 people, resulting in 103 images. 
In each image, the absolute salience of a face was calculated as as ratio of 
the fixation points in the face bounding-box to the total number of fixation points in all 
the face boxes in the image. This results in a ranking of people 
according to their saliency scores. 

\begin{table}[t]
\centering
\begin{tabular}{@{\extracolsep{\fill}}c|ccc@{\extracolsep{\fill}}}
\toprule
&\multicolumn{3}{c}{Importance}\\
Salience &{\small significantly} & {\small slightly} & {\small about} \\
&{\small more} &{\small more} &{\small same} \\
\midrule
{\small significantly-more} & {\small $38.33\%$} & {\small $38.33\%$} & {\small $23.33\%$} \\
{\small slightly-more} & {\small $22.66\%$} & {\small $32.81\%$} & {\small $44.53\%$} \\
{\small  about-same} & {\small $03.82\%$} & {\small $19.51\%$} & {\small $76.67\%$} \\
\bottomrule
\end{tabular}
\caption{Distribution of Importance pair categories among Salience pair categories }
\label{tab:saliency_1}
\vspace{-20pt}
\end{table}

We then collected pairwise importance annotations for this dataset on Mechanical Turk  
using the same interface as used for the the Image-Level Importance dataset. 
Since this dataset is smaller, we annotated all possible face pairs (from the same image). 
Thus, we can extract a full ranking of individuals in each image based on their importance. 
Human judgement-based pairwise annotations are often inconsistent 
(\eg $s_i > s_j$, $s_j > s_k$, and $s_k > s_i$). 
Thus, we used the Elo rating system to obtain a full ranking.

We measured the correlation between importance and saliency rankings using Kendall's Tau. 
The Kendall's Tau was $0.5256$. The most salient face 
was also the most important person in $52.56\%$ of the cases. 

\begin{figure*}[t]
\includegraphics[width=1\textwidth]{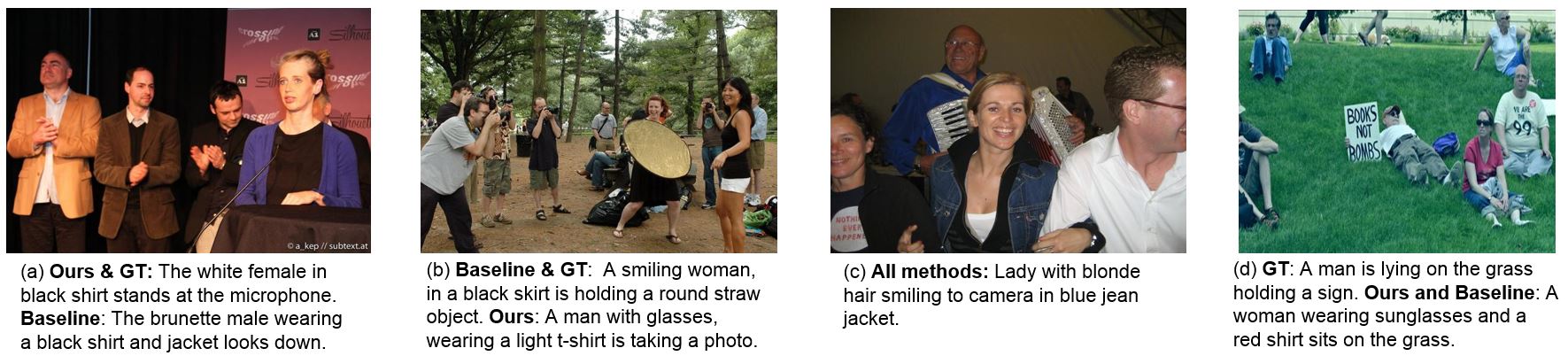}
\caption{Qualitative results for the pruning descriptions experiment}
\label{fig:sentExamples}
\vspace{-15pt}
\end{figure*}

\figref{fig:salExamples} shows qualitative examples of individuals who are judged by humans to be 
salient but not important, important but not salient, both salient and important, and neither. 
Table \ref{tab:saliency_1} shows the `confusion matrix' of saliency \vs importance, 
broken down over the three strength categories. It can be seen that most face-pairs that are `about-same' 
salient are also `about-same' important whereas the other other two categories have less agreement -- 
in a pair $(p_i, p_j)$, 
$p_i$ may be more salient than $p_j$ but less important, 
and vice versa.
\vspace{\sectionReduceTop}
\section{Application: Improving Im2Text}
\label{sec:applications}
\vspace{\sectionReduceBot}


We now show that importance estimation can improve im2text by producing 
more human-like image descriptions, 
as championed by the recent work of Vedantam~\etal~\cite{ramaCider}.  

Sentence generation algorithms~\cite{amirPeople,babytalk} often approach 
the task by first predicting attributes, actions, and other relevant information for every person in an image.
Then these predictions are combined to produce a description for the photo. 
In group photos or crowded scenes, such an algorithm would identify several people 
in the image, and may end up producing overly-lengthy rambling descriptions. 
If the relative importance of the people in the photo is known, the algorithm can focus on the most important people, 
and the rest can be either deemphasized or ignored as appropriate. 
How beneficial is importance prediction in such cases? This experiment addresses this question quantitatively. 

\textbf{Setup.} Our test dataset for this experiment is a set of randomly 
selected 50 images from the Image-Level dataset.  
The training set comprises the remaining 150 images. 
Since the implementation for im2text methods was not available online at the time this work was done, 
we simulated them in the following way. 
First, we collected 1-sentence descriptions for every individual in the test set 
on Mechanical Turk. The annotation interface for these tasks asked 
Turkers to only describe the individual in question. 

\textbf{Prediction.} 
We trained the importance model on the 150 training images and made predictions on the test set. 
We use the predicted importance to find the most important person in the image according to our approach. 
Similarly, we get the most important persons according to the center and random baselines. 
For each selection method, we choose the corresponding 1-sentence description. 
We then performed pair-wise forced-choice tests on Mechanical Turk with these descriptions, asking Turkers 
to evaluate which description was better, and found out the `best' description per image. 



\textbf{Results.}
The importance methods were evaluated by how often their descriptions `won' i.e., 
was ranked as the best description. The results in Table \ref{tab:sentPruning} show that reasoning 
about importance of people in an image helps significantly. Our approach outperformed the `Random' baseline
by 35\%, 
which picks a human-written sentence about a random person in the image. 
An `oracle' that picks the sentence corresponding to the most important person according to the ground-truth 
provides an upper bound (71.43\%) 
on how well we can hope to do 
if we are describing an image with a single sentence about one person. 

\begin{table}[h]
{\small
\centering
\begin{tabular}{@{\extracolsep{\fill}}p{2.5cm}|c}
\toprule
Method & Accuracy  \\
\midrule
Our approach & \textbf{57.14\%} \\
Center & 48.98\%\\
Random & 22.45\%\\
Oracle & \textbf{71.43\%}\\
\bottomrule
\end{tabular}
\caption{Importance prediction improves image descriptions: 
Each row reports the percentage of time the corresponding description was selected as the `best' description. }
\label{tab:sentPruning}
\vspace{-2pt}
}
\end{table}
\vspace{\sectionReduceTop}
\section{Conclusions}
\label{sec:conclusions}
\vspace{\sectionReduceBot}

To summarize, we proposed the task of 
automatically predicting the importance of individuals in group photographs, 
using a variety of features that capture the pose and 
arrangement of the people (but not their identity). 
We formulated two versions of this problem -- 
(a) given a single image, ordering the people in it by relative importance, and 
(b) given a corpus of images for a person, ordering the images by importance of that person. 
We collected two importance datasets to evaluate our approach, and these will be broadly useful to 
others in the vision and multimedia communities. 

Compared to previous work in visual saliency, the proposed person importance is correlated but 
not identical. Saliency is not the same as importance, and saliency predictors cannot be used in the 
place of importance predictors. People in photos may be salient but not important, important but not salient, 
both, and neither. 
Finally, we showed that our method can successfully predict the importance of people from purely visual 
cues, and incorporating predicted importance provides significant improvement in im2text. 

The fact that our model performs close to the inter-human agreement suggests that a more challenging 
dataset should be collected. Compiling such a dataset, with richer attributes such as gender and age, 
and incorporating social relationship and popularity cues 
are the next steps in this line of work.

\newpage
\textbf{Acknowledgements.} 
We thank Amir Sadovnik, Congcong Li, Makarand Tapaswi, and Martin Bauml for assisting 
us in various stages of our work. 
CSM and DB were partially supported by 
the Virginia Tech ICTAS JFC Award, 
the National Science Foundation under grants IIS-1353694 and IIS-1350553, 
the Army Research Office YIP Award W911NF-14-1-0180, 
and the Office of Naval Research grant N00014-14-1-0679. 
The views and conclusions contained herein are those of the authors and should not be 
interpreted as necessarily representing the official policies or endorsements, either expressed or implied, 
of the U.S. Government or any sponsor. 


{\small
\bibliographystyle{ieee}
\bibliography{dbatra,clint}
}

\end{document}